\documentclass{article}
\usepackage[utf8]{inputenc}
\usepackage{graphicx} 
\usepackage{todonotes} 
\PassOptionsToPackage{hyphens}{url}\usepackage{hyperref}
\usepackage{float} 

\usepackage{amsmath} 
\usepackage{booktabs} 
\usepackage{multirow}
\usepackage{tabularx}
\PassOptionsToPackage{numbers,sort&compress}{natbib}
\PassOptionsToPackage{hyphens}{url}

\usepackage[main, final]{neurips_2026}

\setuptodonotes{color=blue!20}

\title{Agentic systems for breast cancer treatment recommendations}

\author{Vinicius Anjos de Almeida \thanks{Vinicius Anjos de Almeida is with Spesia, Alameda Dom Pedro II, 574, 80420-060, Curitiba, Paraná, Brazil, and also with Faculdade de Medicina, Universidade de São Paulo, Av. Dr. Arnaldo, 455, 01246-903, São Paulo, São Paulo, Brazil (corresponding author, e-mail: vinicius.almeida@spesia.com.br).} \and Nícolas Henrique Borges \thanks{Nícolas Henrique Borges is with Spesia (e-mail: nicolas.borges@spesia.com.br).} \and Leonardo Vicenzi \thanks{Leonardo Vicenzi is with Spesia and also with the Laboratory of Artificial Intelligence Applied to Bioinformatics, SEPT, Universidade Federal do Paraná (UFPR), Rua Alcides Vieira Arcoverde, 1225, 81520-260, Curitiba, Paraná, Brazil (e-mail: leonardo.vicenzi@spesia.com.br).} \and Helena Kociolek \thanks{Helena Kociolek is with Spesia, Pontifícia Universidade Católica do Paraná (PUCPR), R. Imac. Conceição, 1155, 80215-901, Curitiba, Paraná, Brazil, and Universidade Federal do Paraná (UFPR), Rua Alcides Vieira Arcoverde, 1225, 81520-260, Curitiba, Paraná, Brazil (e-mail: helenakociolek@hotmail.com).} \and Sarah Miriã de Castro Rocha \thanks{Sarah Miriã de Castro Rocha is with Spesia and also with Pontifícia Universidade Católica do Paraná (PUCPR) (e-mail: sarah.miria@pucpr.edu.br).} \and Frederico Nassif Gomes \thanks{Frederico Nassif Gomes is with Spesia and also with Pontifícia Universidade Católica do Paraná (PUCPR) (e-mail: frednassifgomes@gmail.com).} \and Júlia Cristina Ferreira Ribeiro \thanks{Júlia Cristina Ferreira Ribeiro is with Spesia and also with Pontifícia Universidade Católica do Paraná (PUCPR) (e-mail: f.cristina1@pucpr.edu.br).} \and Lucas Emanuel Silva e Oliveira \thanks{Lucas Emanuel Silva e Oliveira is with Spesia and also with Pontifícia Universidade Católica do Paraná (PUCPR) (e-mail: lucas.oliveira@spesia.com.br).}\\%
}

\begin{document}

\maketitle

\begin{abstract}
    Large language models (LLMs) are increasingly being explored for clinical decision support, but their reliability in complex oncology treatment planning remains unclear. We evaluated agentic LLM systems for breast cancer treatment recommendation generation using 72 real clinical cases across stages I to IV and 1,147 case-specific rubrics generated through Asymmetric Information Rubric Generation (AIRG), in which the rubric generator had access to real clinical decisions unavailable to the evaluated models. Seven pipelines were compared, including single-LLM baselines, tool-augmented systems, and multi-agent architectures with fact checking and autonomous subagent spawning. The best-performing configuration, Claude Opus 4.8 with the D\&C+SA pipeline, achieved a global score of 0.594 ± 0.025. Tool use and increased agent autonomy had mixed effects, improving performance in some settings but degrading it in others. Performance varied by clinical domain and disease stage, and oncologist-led error analysis revealed persistent clinically relevant failures, including incorrect or missing recommendations, flawed justifications, citation errors, outdated claims, and overconfidence. These findings suggest that agentic LLM systems can generate clinically relevant breast cancer recommendations, but remain insufficient for unsupervised clinical use.

\end{abstract}

\section{Introduction}

Large language models (LLMs) have rapidly improved in biomedical and clinical tasks, achieving high performance on medical licensing examinations and other standardized benchmarks \cite{Singhal2023, pricepertokenMedQALeaderboard, Chen2026}. Both general-purpose LLMs and specialized clinical AI systems have demonstrated substantial potential to support healthcare providers in navigating medical knowledge \cite{Vishwanath2026, feng2026expertevaluationclinicalai}. These results have motivated growing interest in the use of LLMs as clinical decision-support tools, particularly in domains where clinicians must synthesize heterogeneous information from guidelines, clinical notes, scientific literature, and patient-specific factors.

Breast cancer treatment is a clinically important setting for evaluating LLM-based decision support. Treatment planning may involve systemic therapy, surgery, radiotherapy, genetic testing, complementary exams, and sequencing decisions that vary according to stage, tumor biology, patient condition, prior treatments, local availability of resources, and evolving guidelines. In many institutions, complex cases are discussed in multidisciplinary tumor boards, where recommendations are shaped by the combined expertise of oncologists, surgeons, radiation oncologists, radiologists, pathologists, geneticists, and other specialists. This makes breast cancer treatment recommendation generation a challenging and clinically meaningful task for evaluating whether LLM systems can support nuanced therapeutic decision-making.

Previous studies have explored the use of LLMs in breast cancer care, including clinical decision support in tumor boards \cite{Sorin2023, Lukac2023, Griewing2023, KARABUA2026, Stalp2024}, surgical planning \cite{COSSU3271}, guideline-based question answering \cite{Rao2023, Haver2023}, and information extraction pipelines \cite{Shankar2026}. Similar applications have also been studied in other cancer types, including colon cancer \cite{KUS2025}, prostate cancer \cite{Tun2025}, head and neck cancer \cite{Lechien2023, Schmidl2024}, and lung cancer \cite{Zabaleta2025}. Although these studies suggest that LLMs may produce clinically useful outputs, they also reveal important limitations. Prior work in breast cancer tumor board settings has reported incomplete agreement with expert recommendations, harmful or inappropriate decisions, poor understanding of treatment sequencing, difficulty integrating surgical history, and only moderate concordance with oncologists in specific treatment scenarios \cite{Sorin2023, Lukac2023, Griewing2023, Griewing2024, KARABUA2026, Stalp2024, ahthiane2025}.

Most previous evaluations have focused on single-model prompting or direct comparisons between LLM-generated recommendations and expert decisions. Less is known about whether agentic systems, tool use, and multi-agent architectures improve treatment recommendation quality in oncology. Liu et al. \cite{Liu2026} developed EvoMDT, a system composed of a multimodal ontology-normalized knowledge base, a five-agent workflow, and a self-improvement loop after each task, in which prompts are optimized based on the system's own feedback about its outputs. The system was evaluated on several question-answering benchmarks and on retrospective data from 405 real patients with lung, breast, or liver cancer, or lymphoma. Each case included a structured summary of the patient's condition and expert treatment plans, which served as the reference for evaluation. EvoMDT achieved response quality comparable to that of physicians, particularly in safety and completeness, and showed stable performance across lymphoma cases. It also provided an advantage in response time compared with clinicians. Other attempts include combining GPT-4 with multimodal precision oncology tools \cite{Ferber2025}, using general medical calculation tools \cite{Goodell2025}, risk-prediction tools \cite{liu2026riskagentsynergizinglanguagemodels}, and domain-specific retrieval pipelines \cite{Ferber2024}. To the best of our knowledge, how models behave when provided with different combinations of tools and different degrees of autonomy remains underexplored.

In principle, access to tools such as web search, PubMed search, and full-text retrieval may allow LLMs to retrieve up-to-date evidence and guidelines. Similarly, multi-agent architectures may decompose complex clinical tasks into more manageable subtasks, such as systemic therapy, surgery, radiotherapy, and complementary exams. However, these approaches also introduce new failure modes. A model must know when to search, how to formulate useful queries, how to assess source relevance, how to integrate retrieved evidence with patient-specific information, and how to avoid overconfident or unsupported conclusions. Therefore, the benefit of tool use and agentic decomposition cannot be assumed and must be evaluated empirically.

A second challenge is evaluation. Clinical treatment recommendations are difficult to score using exact-match labels or simple global judgments because correct answers are often partial, context-dependent, and multidimensional. Rubric-based evaluation provides a flexible alternative by decomposing response quality into multiple criteria, allowing models to receive credit for desirable clinical recommendations while being penalized for unsafe, unsupported, or inappropriate content. This approach has been used in recent medical LLM benchmarks such as HealthBench and HealthBench Professional \cite{arora2025healthbenchevaluatinglargelanguage, hicks2026healthbenchprofessionalevaluatinglarge}. However, expert-authored rubrics are expensive to create, especially in specialized domains such as oncology.

This study addresses four research questions: \textbf{RQ1:} How accurately can current frontier LLMs generate breast cancer treatment recommendations from real clinical cases? \textbf{RQ2:} How do external tools, multi-agent decomposition, fact checking, and increased agent autonomy affect recommendation quality, and do these effects vary across models? \textbf{RQ3:} How does system performance vary across breast cancer stages and clinical domains, including systemic therapy, surgery, radiotherapy, and complementary exams? \textbf{RQ4:} What clinically relevant failure modes persist in the recommendations generated by top-performing agentic systems?

We evaluate agentic systems for breast cancer treatment recommendation generation using real clinical cases and case-specific rubric-based scoring. We make four main contributions. First, we benchmark seven treatment recommendation pipelines, including single-LLM baselines, tool-augmented systems, and multi-agent architectures with increasing agent autonomy, using a curated dataset of 72 breast cancer cases spanning stages I to IV and 1,147 case-specific rubrics grounded in real oncologists' decisions. Second, we introduce AIRG as a pragmatic workflow for generating evaluation rubrics in specialized clinical domains where expert input is scarce or costly. Third, we provide a detailed oncologist-led error analysis of responses generated by a top-performing model-pipeline combination to characterize clinically relevant failure modes. Fourth, we release the source code for our experiments to support reproducibility and future work on clinically grounded evaluation of medical AI systems.

\section{Methods}

\paragraph{Data collection.}The raw data were collected from a private oncology clinic in Curitiba, Brazil. The sampled cases consisted of patients with histologically confirmed invasive breast cancer who attended the institution between 2023 and 2025. Cases were sampled from the electronic health record (EHR) database and were included if, after careful review of the clinical notes, they contained sufficient clinical, pathological, biomarker, staging, and treatment information, and if the clinical decisions made by the oncologist responsible for the case were clearly stated.

In some cases, clinical decisions included recommendations from the institutional multidisciplinary tumor board (MTB), which were also considered a reference for the case. The MTB is a standard multidisciplinary decision-making process used in several health institutions worldwide and, when applicable, may involve clinical oncologists, breast surgeons, radiation oncologists, radiologists, pathologists, geneticists, and other specialists involved in cancer care. MTB sessions are dedicated to complex cases in which clear guidelines or robust scientific evidence are not available, or in which available resources are not ideal and a clinical decision must be made.

After sampling and EHR review, a total of 72 cases were selected to build the evaluation dataset. These cases included 18 patients representing each breast cancer stage: I, II, III, and IV.

A team of five medical students supervised by a certified physician reviewed the clinical notes from each case and created a structured case summary, including demographic data, diagnosis, histopathological diagnosis, biomarkers, and patient performance status scores. Each summary was followed by a set of clinical decisions made for that patient’s care, including systemic therapies, surgical treatments, radiotherapy, and complementary exams ordered.

\paragraph{Toolkit for AI models.} LLMs are often equipped with tools to expand their utility across various tasks. In this study, two tools were used in the form of Model Context Protocol (MCP) servers: web search, powered by Parallel.ai \cite{parallelParallelSystems}; and PubMed search, provided through the official PubMed public API \cite{nihAPIsDevelop}. MCP \cite{githubModelContext} is an open-source project hosted by the Linux Foundation \cite{linuxfoundationLinuxFoundation} that allows AI applications to connect to external data sources, tools, and workflows in a consistent way.

Several companies offer web search tools for agentic systems, including OpenAI \cite{openaiSearchOpenAI}, Google \cite{googleGeminiGoogle}, Exa.ai \cite{exa}, Tavily \cite{tavilyTavily}, and Parallel.ai \cite{parallelParallelSystems}. For this study, the Parallel.ai search tool was selected because of its performance in several published benchmarks and its cost efficiency \cite{parallelaibenchmark}. Comparing different web search tools was beyond the scope of this study.

The National Center for Biotechnology Information provides a public API \cite{nihAPIsDevelop} that allows automated systems to search the PubMed database, retrieve article information and abstracts, and, for open-access publications, retrieve the full manuscript. This API service was converted into an MCP server through an open-source implementation \cite{pubmedmcp}, which was customized for this study.

\paragraph{Rubric generation.} The use of rubrics in this study was directly inspired by the HealthBench \cite{arora2025healthbenchevaluatinglargelanguage} and HealthBench Professional \cite{hicks2026healthbenchprofessionalevaluatinglarge} benchmarks developed by OpenAI. A rubric consists of a criterion and an associated score. The criterion defines an attribute of the text being evaluated that may be present or absent. The score is a value ranging from -10 to +10, indicating the extent to which that attribute is desirable or undesirable in textual responses for the task at hand. Ideally, rubrics should be written directly by domain experts and peer-reviewed to improve consistency. In practice, however, expert availability is often limited and costly, making rubric development a frequent bottleneck in healthcare evaluation tasks.

In this study, we propose \textbf{Asymmetric Information Rubric Generation (AIRG)} as a practical workflow for generating case-specific clinical rubrics when expert input is scarce or expensive. In natural language processing tasks where model outputs must be evaluated along dimensions that cannot be captured by rule-based scoring or straightforward manual review, rubric-based evaluation can use an LLM-as-a-judge to assess each rubric individually and assign scores accordingly. However, rubrics generated only from the input case, without grounding in a gold-standard reference, does not add value for model evaluation because they lack a trustworthy basis for comparison. This workflow addresses this limitation by giving the rubric generator access to privileged information that is not available to the model being evaluated. We hypothesize that this asymmetry allows the generated rubrics to be grounded in reference information and to better reflect clinically meaningful aspects of response quality in the final evaluation scores. In practice, this approach may mitigate the cost of expert rubric creation by using other textual sources of reference information, provided that they are accessible only to the rubric generator.

Prior work has explored rubric-based LLM evaluation, including physician-authored clinical rubrics \cite{arora2025healthbenchevaluatinglargelanguage, hicks2026healthbenchprofessionalevaluatinglarge}, reference-guided LLM judges \cite{kim2024prometheusinducingfinegrainedevaluation}, self-adaptive rubrics \cite{fan-etal-2024-sedareval}, multi-agent rubric generation for dialogue systems \cite{chen2026automatedrubricsreliableevaluation}, grounding on similar previous queries \cite{dhole2026rubricraginterpretablereliablellm}, dataset- and instance-specific rubrics generation with no reference \cite{wang2026generatingrefiningdynamicevaluation}, and automatic extraction of evaluation criteria from gold answers \cite{yu2025pointwisescoresdecomposedcriteriabased}. Our process differs from these approaches by operationalizing this idea in a clinically grounded oncology setting, where the rubric generator has privileged access to real treatment decisions extracted from the EHR and also to literature-search tools, while evaluated systems receive only the case summary. This creates an intentional information asymmetry designed to mitigate the need for fully expert-authored rubrics while preserving reference grounding.

Concretely, a frontier LLM (`gemini-3.5-flash') configured with high reasoning effort was provided with the case summary, the reference clinical decisions made by healthcare providers for that case, and access to web search and PubMed search tools. The model was prompted to generate a set of rubrics for evaluating the quality of treatment recommendations for each case. The prompt instructed the model to cover all relevant aspects of treatment recommendations, including recommended interventions, non-recommended interventions, and treatment alternatives, based on scientific references and official breast cancer guidelines.

Rubrics were also grouped in four categories: systemic therapy, radiotherapy, surgical therapy, and complementary exams.

\begin{figure}[H]
\centering
\includegraphics[width=1.0\textwidth]{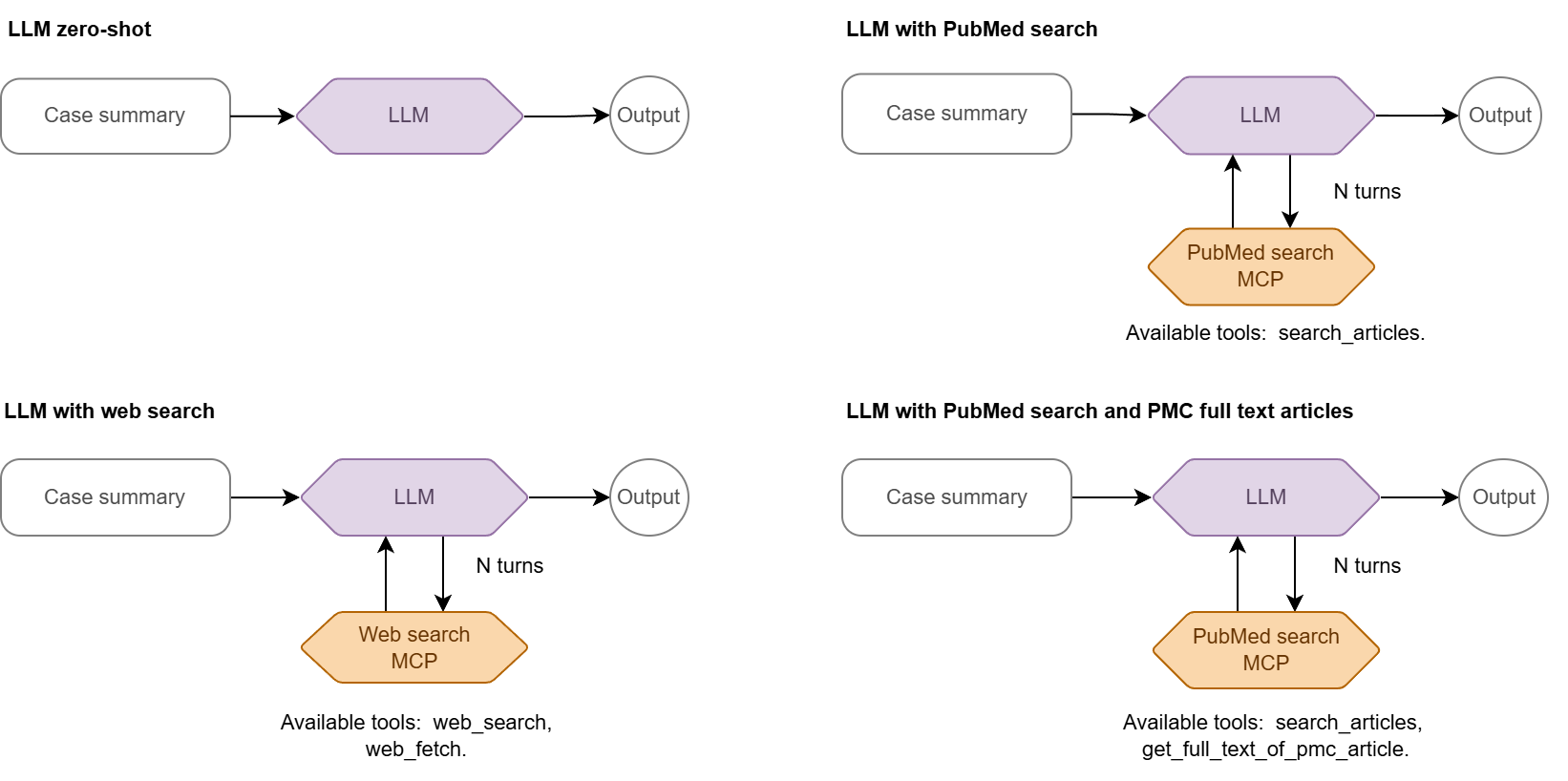}
\caption{The three multi-agent pipelines evaluated for generating treatment recommendations: divide and conquer, divide and conquer with fact checker, and divide and conquer with autonomously spawned subagents and a fact checker. Tools are represented by their keys: web\_search; web\_fetch, for fetching entire webpages; search\_articles, to run queries in the PubMed database and retrieve papers and their abstracts; and get\_full\_text\_of\_pmc\_article, to retrieve a full manuscript from PMC.}
\label{fig:pipelinesdiagram1}
\end{figure}

\begin{figure}[H]
\centering
\includegraphics[width=1.0\textwidth]{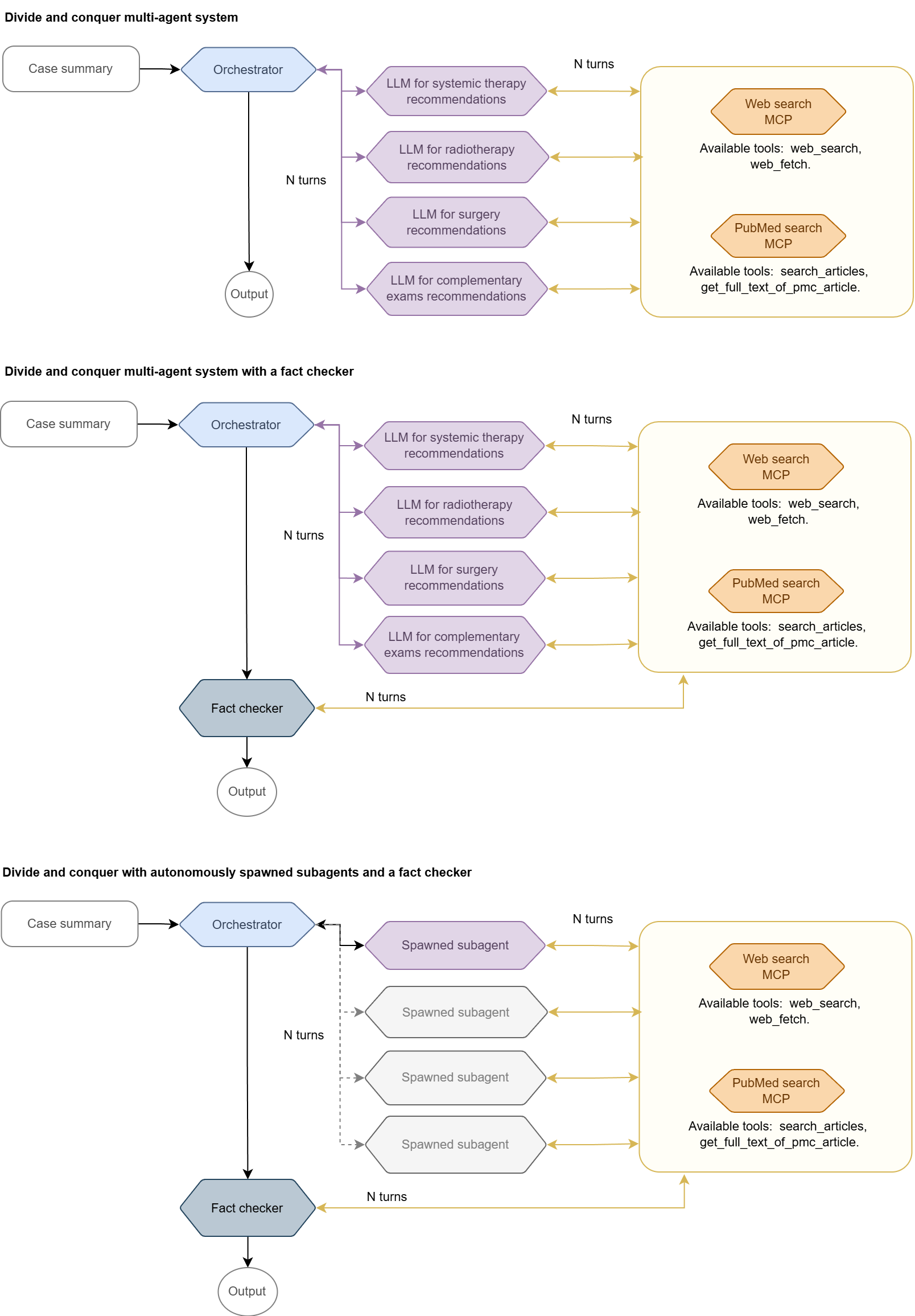}
\caption{The three multi-agent pipelines evaluated for generating treatment recommendations: divide and conquer, divide and conquer with fact checker, and divide and conquer with autonomously spawned subagents and a fact checker. Tools are represented by their keys: web\_search; web\_fetch, for fetching entire webpages; search\_articles, to run queries in the PubMed database and retrieve papers and their abstracts; and get\_full\_text\_of\_pmc\_article, to retrieve a full manuscript from PMC.}
\label{fig:pipelinesdiagram2}
\end{figure}

\paragraph{Treatment recommendation pipelines.} Seven different pipelines for generating treatment recommendations were compared and they are visually represented in \textbf{Figures \ref{fig:pipelinesdiagram1} and \ref{fig:pipelinesdiagram2}} for better understanding. They include:

\begin{itemize}
    \item \textbf{Baseline}: a single LLM with no tools, prompted to generate treatment recommendations from a case summary.
    \item \textbf{LLM with web search (WS)}: a single LLM with access to the web search tool.
    \item \textbf{LLM with PubMed search (PM)}: a single LLM with access to the PubMed search tool.
    \item \textbf{LLM with PubMed search and PMC manuscripts (PM+PMC)}: a single LLM with access to the PubMed search tool and the PMC tool for retrieving the complete manuscripts of open-access articles.
    \item \textbf{Divide and conquer (D\&C)}: a multi-agent architecture comprising one main orchestrator agent and four subagents dedicated to specific topics in breast cancer treatment recommendation generation: systemic therapy, radiotherapy, surgical treatment, and complementary exams. Each subagent had access to web search, PubMed search, and PMC full-text retrieval.
    \item \textbf{Divide and conquer with fact checker (D\&C+FC)}: the same architecture as D\&C, with a handoff to a fact checker agent. This agent receives an initial version of the treatment recommendations and is prompted to check factual claims and identify inconsistencies before generating the final version. The fact checker also has access to all tools but does not have access to previous tool outputs or reasoning traces from the orchestrator agent.
    \item \textbf{Divide and conquer with fact checker and auto-spawning agents (D\&C+SA)}: a multi-agent architecture that, instead of using predefined subagents, can autonomously spawn subagents with customized instructions to generate treatment recommendations. Each subagent has access to the search tools, and the generated recommendations also pass through the fact checker agent before the final output is produced.
\end{itemize}

\paragraph{LLMs evaluated.} Some of the most recent models from major providers were selected for evaluation: gemini-2.5-flash, gemini-3.1-pro-preview, and gemini-3.5-flash from Google; gpt-5.5-2026-04-23 from OpenAI; and claude-opus-4-8 from Anthropic. All models were used with their default reasoning-effort settings. Due to cost constraints, experiments with OpenAI and Anthropic models were limited, but included at least the baseline and the best-performing pipeline identified using Google models.

\paragraph{Metrics.} We used the example-level scoring procedure proposed in HealthBench \cite{arora2025healthbenchevaluatinglargelanguage}. For each example $i \in \{1, \ldots, N\}$, the model-generated answer is evaluated against a set of $M_i$ rubric criteria. Each criterion $j$ has an associated point value $p_{ij} \in [-10, 10]$, where positive values indicate desirable attributes and negative values indicate undesirable attributes. A grader model determines whether each criterion is satisfied by the answer. Let $\mathbf{I}_{ij}$ be an indicator variable equal to 1 when criterion $j$ is satisfied for example $i$, and 0 otherwise. The score for example $i$ is then computed as:

\[
s_i =
\frac{
\displaystyle \sum_{j=1}^{M_i} \mathbf{I}_{ij} p_{ij}
}{
\displaystyle \sum_{j=1}^{M_i} \max(0, p_{ij})
}.
\]

Thus, the numerator represents the total number of points earned by the response, including both positive points for desirable criteria and negative points for undesirable criteria that are present. The denominator corresponds to the maximum attainable positive score for that example. Finally, the overall score $S$ is computed as the clipped mean of the example-level scores:

\[
S = \operatorname{clip}\left(\frac{1}{N}\sum_{i=1}^{N} s_i, 0, 1\right).
\]

This clipping step ensures that the final aggregate score remains bounded between 0 and 1, even though individual example-level scores may be negative when responses satisfy criteria with negative point values. To estimate uncertainty around the aggregate score, we computed the standard deviation of $S$ using nonparametric bootstrapping over examples.

The same procedure was applied separately to each rubric category to estimate its contribution to the final score. For category-specific scores, rubrics were first filtered by category name, and the score was then computed using only the rubrics assigned to that category.

\paragraph{Error analysis.} We conducted a qualitative error analysis of model-generated responses through detailed manual review by one certified oncologist. Each response was examined comprehensively, with errors categorized using both prespecified and emergent categories. Prespecified categories included fabricated references, citation errors, inappropriate or missing clinical recommendations, and incorrect, incomplete, or absent justifications for otherwise appropriate recommendations. Additional error categories were created when necessary during review. The goal of this analysis was to characterize model failures as precisely and comprehensively as possible, capturing both factual and clinical reasoning errors. The specialist also tracked the time needed to complete the task.

\paragraph{Source code.} The source code is available at \url{https://github.com/GRUPOMED4U/breast_cancer_agents_paper}.

\section{Results}

The final evaluation dataset included 72 breast cancer cases, with 18 cases from each clinical stage: stage I, stage II, stage III, and stage IV. Across all cases, the AIRG workflow generated 1,147 case-specific rubrics for evaluating treatment recommendations. The median number of rubrics per case was 15 [IQR 14-18], with a minimum of 11 and a maximum of 27 rubrics per case.

Rubrics were distributed across four clinical categories: systemic therapy, radiotherapy, surgical therapy, and complementary exams. Systemic therapy accounted for 450 (39\%) of all rubrics, radiotherapy for 178 (16\%), surgical therapy for 174 (15\%), and complementary exams for 345 (30\%). The distribution of rubrics across categories is also available in \textbf{Table \ref{tab:rubric_distribution}}.

\begin{table}[H]
\centering
\caption{Distribution of generated rubrics by clinical category.}
\label{tab:rubric_distribution}
\begin{tabular}{lcc}
\hline
\textbf{Rubric category} & \textbf{Number of rubrics} & \textbf{Percentage} \\
\hline
Systemic therapy & 450 & 39\% \\
Radiotherapy & 178 & 16\% \\
Surgical therapy & 174 & 15\% \\
Complementary exams & 345 & 30\% \\
\hline
Total & 1,147 & 100\% \\
\hline
\end{tabular}
\end{table}

\begin{figure}[H]
\centering
\includegraphics[width=1.0\textwidth]{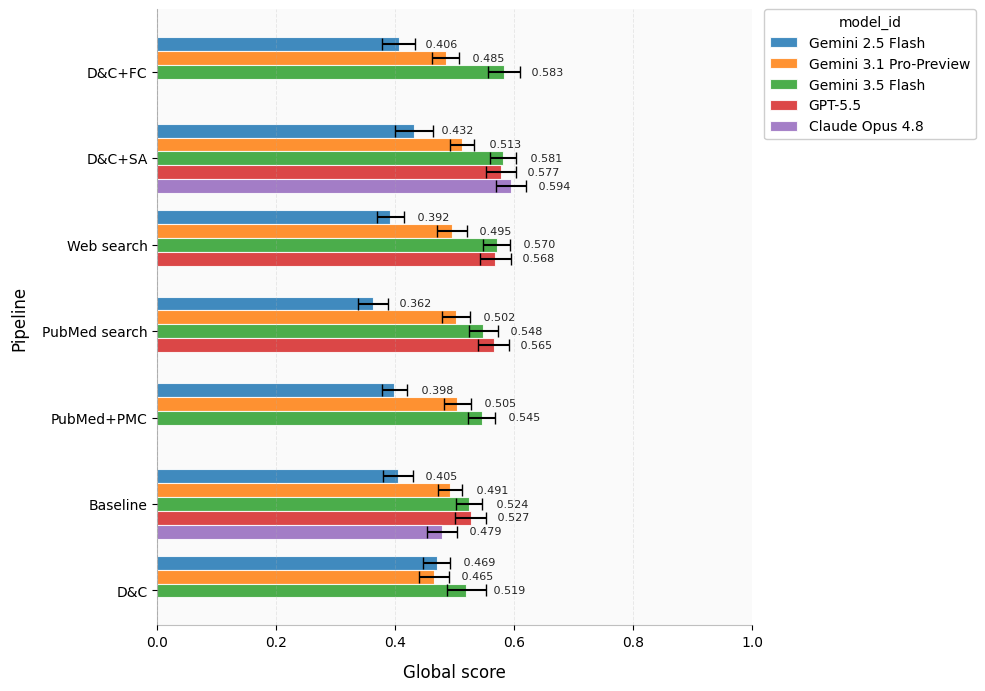}
\caption{Overall benchmark performance across models and treatment recommendation pipelines. Scores were computed using the aggregate rubric-based evaluation procedure, with uncertainty estimated by nonparametric bootstrapping over examples.}
\label{fig:overall_results}
\end{figure}

\begin{figure}[H]
\centering
\includegraphics[width=1.0\textwidth]{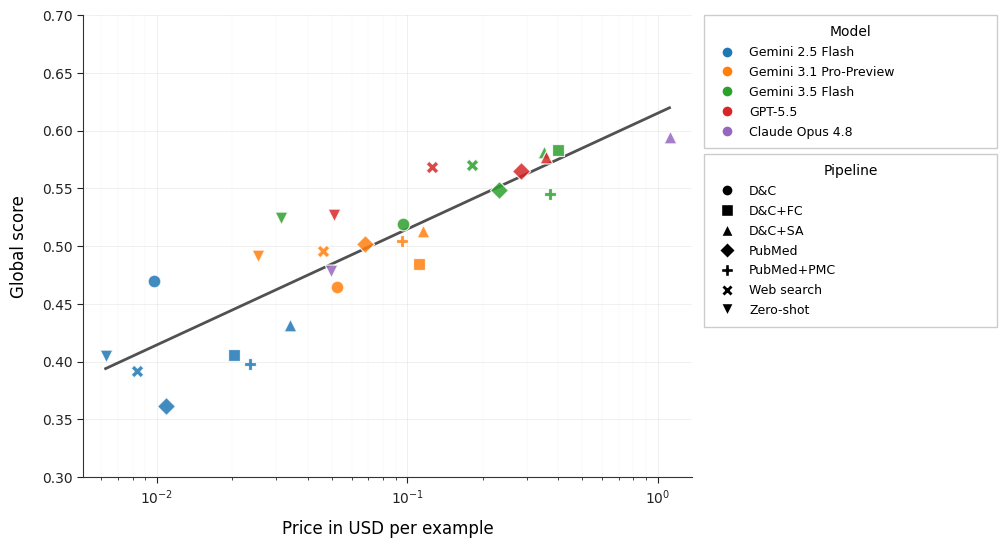}
\caption{Overall benchmark performance across models and treatment recommendation pipelines and their relation with cost per example is USD.}
\label{fig:globalscore_vs_cost}
\end{figure}

\begin{figure}[H]
\centering
\includegraphics[width=1.0\textwidth]{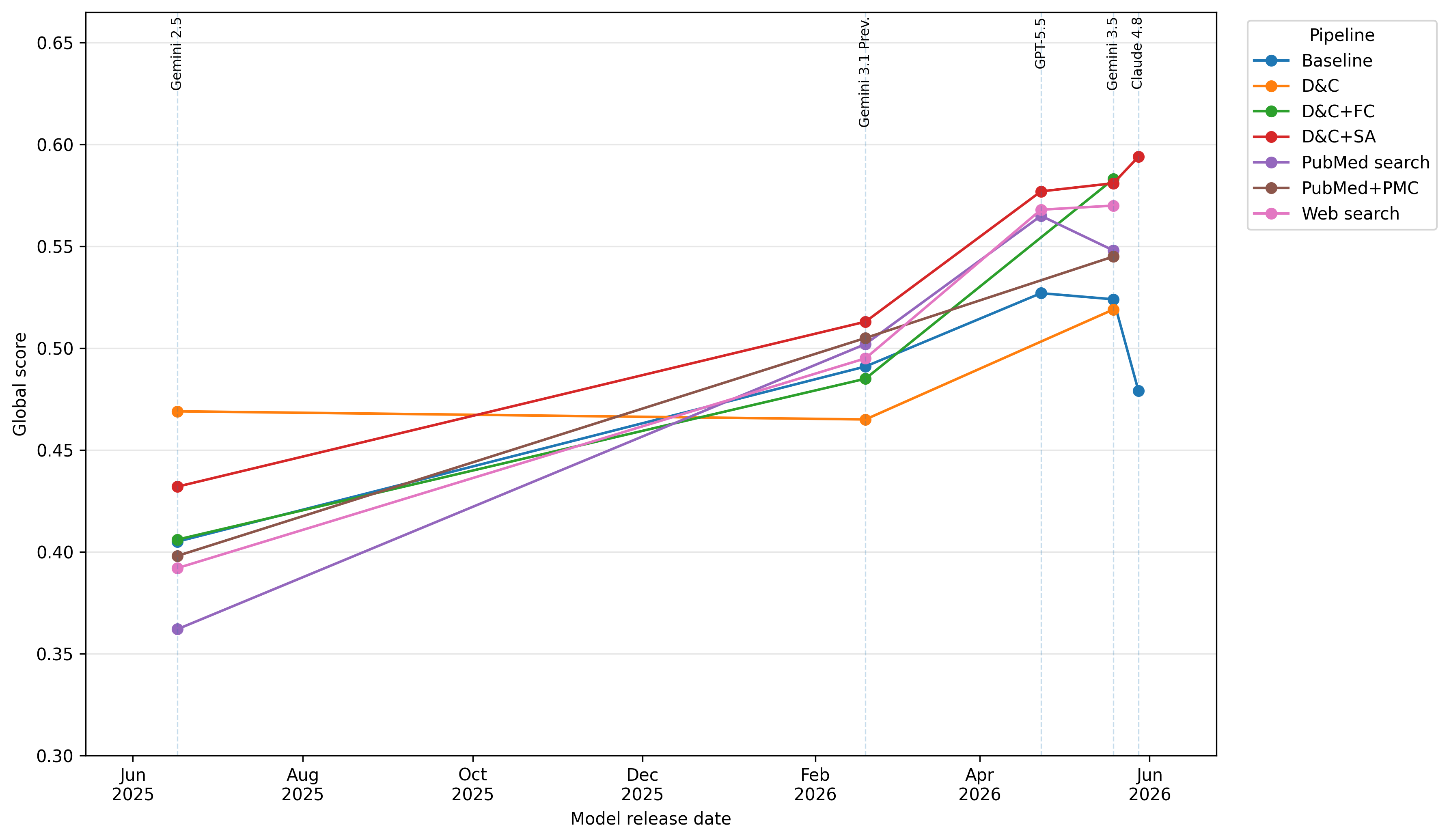}
\caption{Performance across treatment recommendation pipelines, models and their release date.}
\label{fig:scores_by_release_date_vertical_model_labels}
\end{figure}

We evaluated three language models from Google across seven treatment recommendation pipelines, resulting in 21 model-pipeline combinations. In addition, one model from OpenAI was evaluated on baseline, web search, PubMed search and D\&C+SA pipelines; and one from Anthropic was evaluated using the baseline and D\&C+SA pipelines. The overall benchmark scores are shown in \textbf{Figure \ref{fig:overall_results}}. The relationship between cost per task, global score, pipeline, and model is reported in \textbf{Figure \ref{fig:globalscore_vs_cost}}, and their relation with model release date in \textbf{Figure \ref{fig:scores_by_release_date_vertical_model_labels}}. The values used for cost estimation are available in the appendix as \textbf{Table \ref{tab:token_pricing}}.

The highest overall score was achieved by Claude Opus 4.8 using the D\&C+SA pipeline pipeline, with a score of 0.594 $\pm$ 0.025, but the difference between the most recent frontier models is small.

\begin{figure}[H]
\centering
\includegraphics[width=1.0\textwidth]{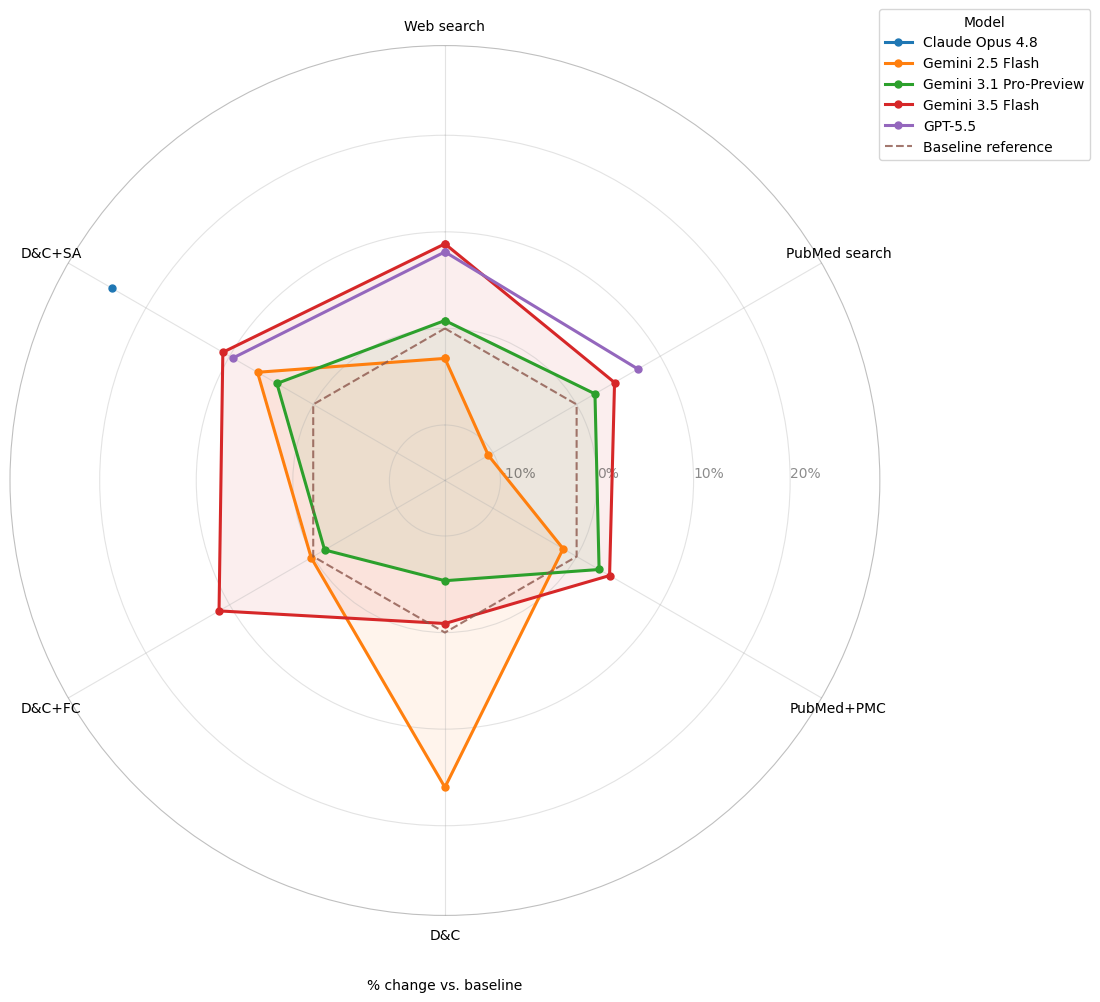}
\caption{Performance gains relative to baseline per pipeline and model.}
\label{fig:pipeline_comparison}
\end{figure}

Pipeline-level performance relative to baseline is summarized in \textbf{Figure \ref{fig:pipeline_comparison}}. Compared with the baseline single-LLM pipeline, the addition of external tools revealed mixed results. The access to tools or the increase in agent autonomy was not necessarily followed by performance gains and varied by model. In some cases, performance deteriorated after providing access to the web search and PubMed search tools (see gemini-2.5-flash results).

\begin{figure}[H]
\centering
\includegraphics[width=1.0\textwidth]{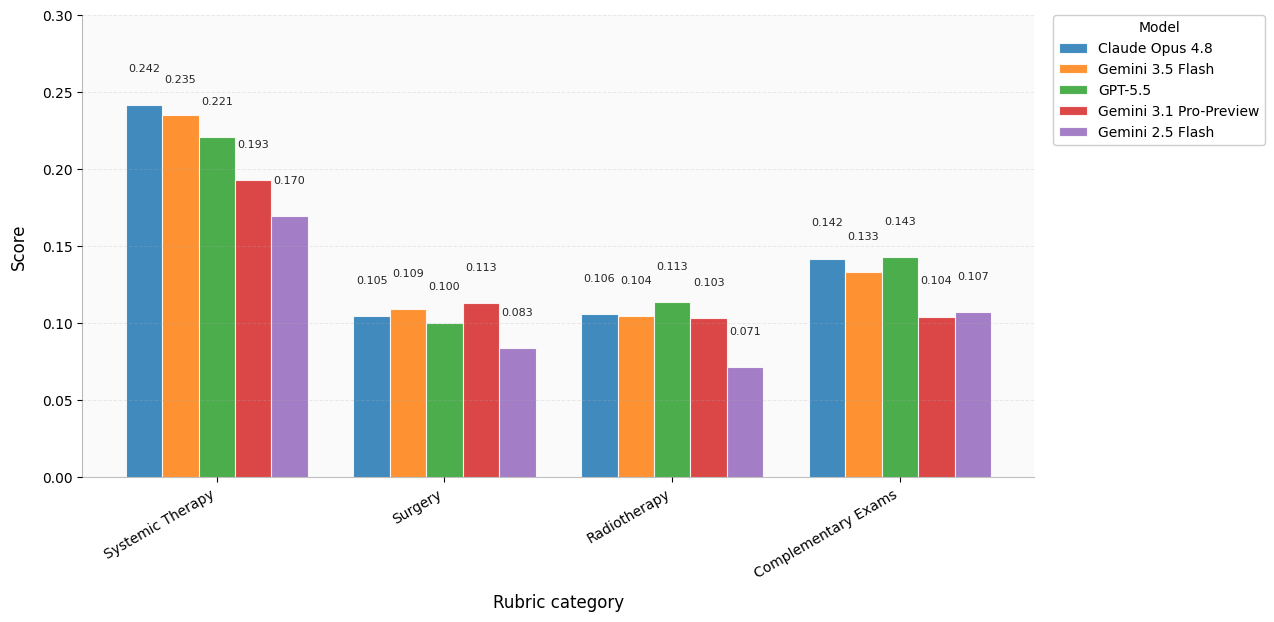}
\caption{Performance gains relative to baseline per pipeline and model.}
\label{fig:decomposed-scores-per-rubric-category}
\end{figure}

To examine which clinical domains contributed most to the final scores, we computed category-specific scores for systemic therapy, radiotherapy, surgical therapy, and complementary exams. We provide a plot with decomposed scores related to D\&A+SA, our best performing pipeline, in \textbf{Figure \ref{fig:decomposed-scores-per-rubric-category}} and a table including results from all experiments is provided in the supplementary material as \textbf{Table \ref{tab:decomposed_scores}}. Higher scores were concentrated in systemic therapy related rubrics, which is also the biggest in total rubric count. The absolute differences between model scores were also higher in the systemic therapy category, with values ranging from 0.170 to 0.242.

We also evaluated performance separately by breast cancer stage. Results are shown in \textbf{Table \ref{tab:stage_results}}. Across all models and pipelines, mean performance was highest for stage I, with a score of 0.545, and lowest for stage III, with a score of 0.441.

Performance differences by stage may reflect variation in clinical complexity, the degree of guideline standardization, and the number of acceptable treatment alternatives. In particular, stages II and III cases showed lower performance, which may be consistent with the greater diversity of breast cancer treatment decisions.

\begin{table}[H]
\centering
\caption{Benchmark scores stratified by breast cancer stage.}
\label{tab:stage_results}
\begin{tabular}{lll}
\hline
\textbf{Stage} & \textbf{Number of cases} & \textbf{Mean score} \\
\hline
I & 18 & 0.545 $\pm$ 0.064 \\
II & 18 & 0.481 $\pm$ 0.075 \\
III & 18 & 0.441 $\pm$ 0.076 \\
IV & 18 & 0.534 $\pm$ 0.070 \\
\hline
\end{tabular}
\end{table}

A detailed error analysis was performed by a certified oncologist over 10 responses generated by gemini-3.5-flash using the D\&C+SA pipeline. This model-pipeline combination was selected because it was among the top-performing approaches in the quantitative benchmark. The objective of the error analysis was to verify every claim made by the generative model, identify, quantify and group errors.

The error analysis identified 8 recurrent failure modes. These included: incorrect or incomplete justification; incorrect or missing clinical recommendations; citation errors and reference fabrication; overconfident statements; and outdated clinical claims. Error count and small description of each failure mode are shown in \textbf{Table \ref{tab:error_analysis}}. The oncologist took 15 hours and 45 minutes to review the clinical recommendations generated for the 10 sampled cases, 1 hour and 35 minutes per case on average, ranging from 45 minutes to 2 hours and 30 minutes spent in a single case depending on complexity.

\begin{table}[htbp]
\centering
\caption{Frequency of error types identified in the error analysis. Numbers refer to errors found in a sample of 10 clinical recommendations generated by gemini-3.5-flash in the D\&C+SA pipeline with cases representing all four breast cancer stages.}
\label{tab:error_analysis}

\begin{tabularx}{\textwidth}{%
>{\raggedright\arraybackslash}p{3.5cm}
rr
>{\raggedright\arraybackslash}X
}
\toprule
\textbf{Error type} & \textbf{Count} & \textbf{\%} & \textbf{Description} \\
\midrule
Incorrect justification & 17 & 22.7 & The clinical recommendation was accurate, but the rationale behind it was not. \\
Incorrect clinical recommendation & 15 & 20.0 & Wrong or imprecise recommendation. \\
Citation error & 12 & 16.0 & The reference exists, but the claims made related to the citation are not correct. \\
Missing clinical recommendation & 12 & 16.0 & A relevant clinical recommendation for the case was not included. \\
Overconfident clinical statement & 8 & 10.7 & Expressed overconfidence while the scientific evidence shows uncertainty or the case summary misses important information to support it. \\
Incomplete justification & 5 & 6.7 & Justification was partially correct, but missed important information. \\
Outdated clinical claim & 4 & 5.3 & Answer includes an accurate claim, although outdated for current standards. \\
Reference fabrication & 2 & 2.7 & Mentions documents that do not exist. \\
\bottomrule
\end{tabularx}
\end{table}

\section{Discussion}

In this study, we evaluated agentic systems for generating breast cancer treatment recommendations using a clinically grounded benchmark derived from real cancer cases. Overall, our findings suggest that recent frontier language models can produce clinically relevant treatment recommendations, but that their performance remains far from sufficient for clinical use without careful review of every recommendation. The best-performing model-pipeline combination achieved a global score of 0.594, indicating that even the strongest evaluated systems failed to satisfy a substantial proportion of case-specific clinical criteria.

Previous work has shown that, in a relatively short time, LLMs have improved substantially, achieved very high scores on medical licensing examinations \cite{Singhal2023, pricepertokenMedQALeaderboard}, and reached early-career physician performance \cite{Chen2026}. In some cases, general-purpose LLMs even surpass specialized clinical AI tools on medical benchmarks \cite{Vishwanath2026}. However, their ability to provide medical advice to the general public may not translate into similarly reliable performance \cite{Bean2026, Draelos2026}, and the fabrication of medical facts still imposes a serious risk in clinical practice \cite{Chen2026}. In challenging medical tasks, although some models may outperform physicians in answer accuracy, their reasoning may not align with their final answers, especially in tasks involving image comprehension \cite{Jin2024}. Models can also be highly sensitive to simple adversarial transformations, such as prompt variations, which may negatively affect response quality, even in models specifically trained for medical tasks \cite{alkaeed2026patientdifferentwordsdifferent, Gu2026}. Moreover, performance can remain remarkably unaffected by the removal of key information from the prompt, highlighting the presence of shortcuts and flawed associations in model predictions \cite{Gu2026}.

The heterogeneity observed in studies involving LLMs naturally persists in multi-agent setups. A central finding of this study was that increased system complexity did not uniformly translate into better performance. Although the highest overall score was obtained with the D\&C + SA pipeline, which combines autonomous subagent spawning with a fact-checking step, the benefits of tools and multi-agent architectures varied considerably across models. In some cases, adding web search, PubMed search, or multi-agent decomposition improved performance; in others, it produced little benefit or even degraded performance. This suggests that agentic architecture should not be treated as a universally beneficial design choice, but rather as a model-dependent and task-dependent intervention.

The mixed effect of tool use is particularly important. External tools are often assumed to improve clinical reasoning by giving models access to up-to-date evidence, scientific literature, and guideline-related information \cite{Goodell2025}. However, our results suggest that access to information alone is not enough. Models must also be able to formulate appropriate queries, select relevant sources, interpret retrieved evidence correctly, and integrate that evidence into patient-specific recommendations.

This may explain why weaker models appeared to benefit less from tool access or, in some cases, performed worse after tools were introduced. Tool use increases the cognitive and procedural complexity of the task: the model must decide when to search, what to search for, which retrieved information is trustworthy, and how to reconcile external evidence with the details of the case. As shown in \textbf{Figure \ref{fig:scores_by_release_date_vertical_model_labels}}, although the number of datapoints is small, performance seems to increase as newer models are released. In our setup, newer models also appeared more capable of autonomously interacting with tools and coordinating subagents. This finding favors pipelines that provide more autonomy to agents rather than predefining roles and information flow. It is also aligned with findings suggesting that LLMs have become increasingly capable of handling longer software tasks over time \cite{kwa2026measuringaiabilitycomplete, metr-2026-time-horizons}.

The results also suggest that the optimal architecture may depend on the base model. More capable models may be able to exploit complex multi-agent workflows, whereas less capable models may perform better with simpler prompting strategies. This has practical implications for the design of clinical AI systems: architecture should be selected empirically for each model and task, rather than assumed to generalize across model families.

\subsection{Rubric-based evaluation and AIRG}

A second contribution of this work is the use of Asymmetric Information Rubric Generation (AIRG) to generate case-specific evaluation criteria. In clinical domains, evaluation is difficult because correct answers are often nuanced, context-dependent, and not reducible to exact string matching or simple classification labels. Rubric-based evaluation provides a more flexible framework because it can reward partially correct recommendations, penalize unsafe or inappropriate recommendations, and evaluate multiple dimensions of response quality.

Several groups have explored automatic rubric generation pipelines related to AIRG. Kim et al. \cite{kim2024prometheusinducingfinegrainedevaluation} developed an automatically generated rubric dataset designed to contain diverse and fine-grained assessment rubrics representing user demands. Although their focus was on inducing LLM evaluation capabilities during training, they used 50 sampled seed rubrics that were augmented and incorporated into the training data. Fan et al. \cite{fan-etal-2024-sedareval} used human-made rubrics to train a model to generate rubrics more closely aligned with human judgment, while also highlighting GPT-4 as a cheaper automatic generation option. Chen et al. \cite{chen2026automatedrubricsreliableevaluation} introduced an agentic system with access to domain-specific knowledge bases and verification loops to generate rubrics for medical dialogue systems. Dhole and Agichtein \cite{dhole2026rubricraginterpretablereliablellm} also incorporated retrieval-augmented generation into the rubric generation pipeline. Wang and Blanco \cite{wang2026generatingrefiningdynamicevaluation} proposed a strategy that involves neither reference answers nor human-authored rubrics. Yu et al. \cite{yu2025pointwisescoresdecomposedcriteriabased} started from expert-authored gold answers and their factual elements to generate instance-specific criteria.

The key idea behind AIRG is that the rubric generator has access to privileged reference information that is not available to the evaluated model. This approach also allows evaluation pipelines to scale using data that are already available in electronic health records. In this study, the rubric generator was a frontier model (gemini-3.5-flash) with high reasoning effort and received the case summary, the actual clinical decisions, and access to search tools. In contrast, the evaluated models received only the case summary and the pipeline-specific resources under evaluation. This asymmetry is intended to make the rubrics more clinically grounded than rubrics generated from the input case alone.

However, AIRG should be interpreted as a pragmatic evaluation strategy rather than a replacement for expert-authored rubrics. Automatically generated rubrics may omit important criteria, include criteria of limited clinical relevance, assign inappropriate weights, or encode errors derived from the reference decision or the rubric-generating model. Future work should directly validate AIRG against physician-authored rubrics, measure agreement between human experts and LLM-generated criteria, and test whether AIRG scores correlate with expert global judgments of recommendation quality.

\subsection{Clinical-domain and stage-specific performance}

The category-specific results suggest that model performance was not uniform across clinical domains. Scores were higher for systemic therapy-related rubrics, while performance in surgery, radiotherapy, and complementary exams appeared more variable. This may reflect differences in the amount of available evidence, the degree of guideline standardization, the way recommendations are documented in clinical notes, or the relative frequency of these topics in model pretraining and instruction-tuning data.

Performance also differed across breast cancer stages. Stage I cases showed the highest mean performance, while stage III cases showed the lowest. This pattern is clinically plausible because earlier-stage breast cancer often involves more standardized decision pathways, whereas stage II and especially stage III cases may require more complex integration of tumor biology, local disease extent, neoadjuvant treatment decisions, surgical planning, radiotherapy indications, genetic testing, and patient-specific factors.

These findings align with previously published results in which intermediate-risk profiles and tumor grades had higher discordance rates between clinicians and LLM generated recommendations \cite{KARABUA2026, Stalp2024}. Future benchmarks should not report only aggregate performance. Stratification by clinical stage, treatment domain, and case complexity may reveal important weaknesses that are hidden by global scores. For clinical deployment, such stratified evaluation is essential because average performance may obscure poor performance in precisely the cases where decision support would be most valuable.

\subsection{Error analysis and clinical safety}

The qualitative error analysis provides an important complement to the quantitative benchmark. Even among high-performing model-pipeline combinations, the oncologist review identified multiple clinically relevant failure modes, including incorrect recommendations, missing recommendations, incorrect or incomplete justifications, citation errors, reference fabrication, outdated claims, and overconfident statements.

This finding highlights a major limitation of relying only on aggregate scores. A model may achieve a relatively high benchmark score while still producing errors that would be unacceptable in clinical practice. In particular, overconfident statements and incorrect justifications are concerning because they may make an answer appear more reliable than it is. Citation errors are also important because they may create a false impression of evidence-based reasoning even when the cited source does not support the claim.

The time required for manual error analysis also illustrates the difficulty of evaluating clinical generative AI systems. The oncologist spent 15 hours and 45
minutes reviewing 10 responses, corresponding to approximately 1 hour and 35 minutes per case. This reinforces the need for scalable evaluation methods, but it also shows why fully automated evaluation should be periodically audited by specialists. A practical evaluation workflow may require both broad automated benchmarking and smaller, high-resolution expert error analyses. Critical errors and long reviewing time may hinder practical utility as healthcare providers rarely can dedicate such time and might prefer to manually search on trusted sources. 

As models improve and their responses become more elaborate, identifying gaps and errors becomes increasingly difficult, effortful, and expensive, especially in high-stakes domains such as healthcare. It is essential for the research community to define evaluation strategies that are reliable, scalable, and capable of testing systems with rapidly evolving capabilities. Marro et al. \cite{marro2026benchmarkingedgecomprehension} proposed a strategy that frames evaluation as an adversarial game, in which correctness is understood as resistance to falsification rather than agreement with a trusted source. In the mathematical domain, they demonstrated that a much less powerful model, GPT-3.5, was able to correctly benchmark GPT-5.2. In medicine, applying this idea is particularly challenging because there is often no predefined logical structure for distinguishing true from false. Nevertheless, this remains an open research direction worth pursuing.

\subsection{Implications for clinical AI system design}

Our results have several implications for the development of LLM-based clinical decision-support systems. First, model capability remains a major determinant of performance. Architectural improvements such as tool use, decomposition, fact checking, and autonomous subagent creation may help, but they do not compensate reliably for weaker base models.

Second, retrieval and tool-use systems should be evaluated as complete pipelines, not as isolated components. A system with access to PubMed or web search may still fail if the model retrieves irrelevant evidence, misinterprets the literature, or applies evidence inappropriately to the case.

Third, fact-checking agents should not be assumed to eliminate errors. In this study, fact-checking was associated with performance gains in some settings, but errors persisted even in outputs generated by pipelines that included a fact-checking step. Future work should investigate whether fact-checkers need access to the full reasoning trace, previous tool outputs, or structured intermediate claims to be more effective.

Fourth, multi-agent autonomy may be most useful when the task can be decomposed into clinically meaningful subtasks. Breast cancer treatment naturally involves systemic therapy, surgery, radiotherapy, and complementary exams, making it a favorable setting for divide-and-conquer architectures. Whether similar gains would occur in less decomposable clinical tasks remains uncertain.

\subsection{Strengths and limitations}

A major strength of this study is the use of real breast cancer cases from clinical practice, including cases across all four disease stages. Another strength is the evaluation of multiple models and multiple pipeline architectures, which allowed us to assess not only model performance but also the interaction between model choice and system design. The combination of quantitative rubric-based scoring with qualitative oncologist-led error analysis provides a more complete view of system behavior than either method alone.

This study also has several limitations. First, the dataset was derived from a single private oncology clinic, which may limit generalizability to other institutions, countries, resource settings, and clinical practice patterns. Second, the reference decisions were based on real clinical decisions rather than independent consensus review by multiple oncologists. Although this increases clinical realism, it also means that the reference standard may reflect local practice and individual decision-making.

Third, the rubrics were automatically generated. Although AIRG was designed to ground rubrics in reference information, the generated criteria may contain omissions, inappropriate weights, or model-introduced biases. Fourth, the LLM-as-a-judge evaluation itself may introduce errors, especially in nuanced clinical cases. Fifth, experiments with some frontier models were limited by cost, preventing a fully balanced comparison across all models and pipelines. Sixth, no adversarial transformations were performed to assess the robustness of model generations. We hypothesize that scores would also be vulnerable to such transformations, as changes in prompt wording, formatting, missing information, or distractor information may substantially affect model outputs. Finally, the study did not systematically evaluate different reasoning-effort settings, prompt variants, tool configurations, or fact-checker designs.

\subsection{Future work}

Future studies should validate AIRG in other clinical domains and compare its outputs against expert-authored rubrics. One possible validation strategy would be to apply AIRG to datasets with existing human-written rubrics or ideal completions, then compare automatically generated rubrics with the original expert criteria and assess agreement in final model rankings.

Further work should also investigate how to improve tool use in clinical agentic systems. This includes better query generation, source selection, evidence ranking, guideline retrieval, structured citation verification, and explicit separation between evidence extraction and recommendation synthesis. Another important direction is to evaluate whether fact-checking agents become more effective when they receive structured intermediate claims, retrieved sources, or the outputs of previous subagents.

Finally, future benchmarks should include prospective clinical review, multi-specialist adjudication, and assessment of potential harm. Since clinical recommendations can be partially correct yet still unsafe, evaluation frameworks should explicitly measure not only factual correctness but also clinical appropriateness, uncertainty calibration, omission of necessary care, and risk of misleading justification.

\subsection{Conclusion}

In summary, this study shows that agentic LLM systems can generate clinically relevant breast cancer treatment recommendations, but their performance remains inconsistent and dependent on both the base model and system architecture. Tool use and multi-agent workflows can improve performance in some settings, but they are not universally beneficial and may introduce additional failure modes. AIRG offers a practical approach for scalable rubric-based evaluation in specialized clinical domains, but it requires further validation. Overall, these findings support the continued development of clinically grounded benchmarks, hybrid automated-human evaluation workflows, and safer architectures for medical AI decision support.

\section*{Author contributions}

V.A.A. was responsible for conceptualization, formal analysis, investigation, methodology, software, validation, visualization, and writing - original draft preparation. N.H.B. contributed with conceptualization and data curation. H.K., S.M.C.R., F.N.G., and J.C.F.R. contributed with data curation. L.E.S.O. contributed with funding acquisition and supervision. All authors contributed to final review and editing. 

\section*{Ethical statement}

The study was approved by the institutional review board/ethics committee under protocol CAAE 90422925.6.0000.0096 of Hospital das Clínicas, Universidade Federal do Paraná, Curitiba - Paraná. All the procedures in this study were in accordance with the 1975 Helsinki Declaration, updated in 2013. In addition, the study complied with the Data Use Consent Term, ensuring that all patient data were accessed, handled, and analyzed exclusively for scientific purposes under strict confidentiality and data protection requirements.

\bibliographystyle{vancouver}
\bibliography{references}

\newpage

\appendix

\renewcommand{\thefigure}{A.\arabic{figure}}
\setcounter{figure}{0}
\renewcommand{\thetable}{A.\arabic{table}}
\setcounter{table}{0}

\section{Supplementary material}

\begin{table}[htbp]
\centering
\caption{Token pricing used for cost estimation. Prices are reported in US dollars per one million tokens.}
\label{tab:token_pricing}
\begin{tabular}{lrrrr}
\toprule
\textbf{Model} & 
\textbf{Input} & 
\textbf{Cached input} & 
\textbf{Output} \\
\midrule
Gemini 3.5 Flash        & 1.50 & 0.15 & 9.00  \\
Gemini 2.5 Flash        & 0.30 & 0.30 & 2.50  \\
Gemini 3.1 Pro Preview  & 2.00 & 0.20 & 12.00 \\
GPT-5.5                 & 5.00 & 0.50 & 30.00 \\
Claude Opus 4.8         & 5.00 & 0.50 & 25.00 \\
\bottomrule
\end{tabular}
\end{table}

\newpage

\begin{table}
\centering
\caption{Token pricing used for cost estimation. Prices are reported in US dollars per one million tokens. Systemic t.: systemic therapy; radiot.: radiotherapy; compl. ex.: complemmentary exams.}
\vspace{0.2cm}
\label{tab:decomposed_scores}
\begin{tabular}{llrrrr}
\toprule
 &  & Systemic T. & Surgery & Radiot. & Compl. Ex. \\
Pipeline & Model &  &  &  &  \\
\midrule
\multirow[t]{3}{*}{D\&C} & Gemini 3.5 Flash & 0.213 & 0.084 & 0.096 & 0.126 \\
 & Gemini 3.1 Pro Preview & 0.178 & 0.104 & 0.092 & 0.091 \\
 & Gemini 2.5 Flash & 0.164 & 0.102 & 0.104 & 0.099\\
 \\
\multirow[t]{3}{*}{D\&C+FC} & Gemini 3.5 Flash & 0.232 & 0.104 & 0.102 & 0.146 \\
 & Gemini 3.1 Pro Preview & 0.177 & 0.107 & 0.091 & 0.111 \\
 & Gemini 2.5 Flash & 0.147 & 0.089 & 0.091 & 0.078 \\
\\
\multirow[t]{5}{*}{D\&C+SA} & Claude Opus 4.8 & 0.242 & 0.105 & 0.106 & 0.142 \\
 & Gemini 3.5 Flash & 0.235 & 0.109 & 0.104 & 0.133 \\
 & GPT-5.5 & 0.221 & 0.100 & 0.113 & 0.143 \\
 & Gemini 3.1 Pro Preview & 0.193 & 0.113 & 0.103 & 0.104 \\
 & Gemini 2.5 Flash & 0.170 & 0.083 & 0.071 & 0.107 \\
\\
\multirow[t]{4}{*}{PubMed} & GPT-5.5 & 0.229 & 0.098 & 0.112 & 0.125 \\
 & Gemini 3.5 Flash & 0.217 & 0.112 & 0.102 & 0.117 \\
 & Gemini 3.1 Pro Preview & 0.195 & 0.110 & 0.106 & 0.090 \\
 & Gemini 2.5 Flash & 0.116 & 0.089 & 0.084 & 0.073 \\
\\
\multirow[t]{3}{*}{PubMed + PMC} & Gemini 3.5 Flash & 0.223 & 0.106 & 0.101 & 0.115 \\
 & Gemini 3.1 Pro Preview & 0.204 & 0.104 & 0.104 & 0.092 \\
 & Gemini 2.5 Flash & 0.138 & 0.104 & 0.089 & 0.068 \\
\\
\multirow[t]{4}{*}{Web search} & Gemini 3.5 Flash & 0.227 & 0.105 & 0.103 & 0.135 \\
 & GPT-5.5 & 0.216 & 0.111 & 0.115 & 0.127 \\
 & Gemini 3.1 Pro Preview & 0.193 & 0.101 & 0.102 & 0.099 \\
 & Gemini 2.5 Flash & 0.127 & 0.098 & 0.104 & 0.064 \\
\\
\multirow[t]{5}{*}{Baseline} & Claude Opus 4.8 & 0.227 & 0.136 & 0.182 & 0.000 \\
 & Gemini 3.5 Flash & 0.215 & 0.117 & 0.098 & 0.094 \\
 & GPT-5.5 & 0.206 & 0.103 & 0.118 & 0.100 \\
 & Gemini 3.1 Pro Preview & 0.197 & 0.103 & 0.109 & 0.083 \\
 & Gemini 2.5 Flash & 0.123 & 0.103 & 0.097 & 0.082 \\
\cline{1-6}
\bottomrule
\end{tabular}
\end{table}

\end{document}